\documentclass{llncs}
\usepackage{llncsdoc}
\usepackage{bm}
\usepackage{amsfonts,amssymb,amsmath}
\usepackage[colorlinks=true,citecolor=blue]{hyperref}
\usepackage{graphicx,subfigure}

\begin{document}

\frontmatter          
\pagestyle{headings}  
\mainmatter   

\title{Learning Myelin Content in Multiple Sclerosis from Multimodal MRI through \texorpdfstring{\\ Adversarial Training}{}}
%

\author{Wen Wei\inst{1,2,3} \and Emilie Poirion\inst{3} \and Benedetta Bodini\inst{3} \and Stanley Durrleman\inst{2,3}\\ Nicholas Ayache\inst{1}    \and Bruno Stankoff\inst{3}  \and Olivier Colliot\inst{2,3} }

\authorrunning{Wen Wei et al.} 

\institute{Epione project-team, Inria, Sophia Antipolis, France
\and
Aramis project-team, Inria, Paris, France
\and
Institut du Cerveau et de la Moelle \'epini\`ere, ICM, Inserm U 1127, \\
CNRS UMR 7225, Sorbonne Universit\'e, F-75013, Paris, France
}


\maketitle              

\begin{abstract}
Multiple sclerosis (MS) is a demyelinating disease of the central nervous system (CNS). A reliable measure of the tissue myelin content is therefore essential to understand the physiopathology of MS, track progression and assess treatment efficacy. Positron emission tomography (PET) with $[^{11} \mbox{C}] \mbox{PIB}$ has been proposed as a promising biomarker for measuring myelin content changes in-vivo in MS.
However, PET imaging is expensive and invasive due to the injection of a radioactive tracer. On the contrary, magnetic resonance imaging (MRI) is a non-invasive, widely available technique, but existing MRI sequences do not provide, to date, a reliable, specific, or direct marker of either demyelination or remyelination. In this work, we therefore propose Sketcher-Refiner Generative Adversarial Networks (GANs) with specifically designed adversarial loss functions to predict the PET-derived myelin content map from a combination of MRI modalities. The prediction problem is solved by a sketch-refinement process in which the sketcher generates the preliminary anatomical and physiological information and the refiner refines and generates images reflecting the tissue myelin content in the human brain. We evaluated the ability of our method to predict myelin content at both global and voxel-wise levels. The evaluation results show that the demyelination in lesion regions and myelin content in normal-appearing white matter (NAWM) can be well predicted by our method. The method has the potential to become a useful tool for clinical management of patients with MS.



\end{abstract}





\vspace{-3em}

\section{Introduction}

Multiple Sclerosis (MS) is the most common neurological disability in young adults \cite{pmid18970977}. MS pathophysiology predominately involves autoimmune aggression of central nervous system (CNS) myelin sheaths, which results in inflammatory demyelinating lesions. These demyelinating lesions in CNS can cause different symptoms such as sensory, cognitive or motor skill dysfunctions \cite{pmid18970977}. In MS, spontaneous myelin repair occurs, allowing restoration of secure and rapid conduction and protecting axons from degeneration. However, remyelination is generally insufficient to prevent irreversible disability. Therefore, a reliable measure of the tissue myelin content is essential to understand MS physiopathology, and also to quantify the effects of new promyelinating therapies.



Positron emission tomography (PET) with $[^{11} \mbox{C}] \mbox{PIB}$ has been proposed as a promising biomarker for measuring myelin content changes in-vivo in MS \cite{ANA22320}. In particular, it has been recently described that $[^{11} \mbox{C}] \mbox{PIB}$ PET can be used to visualize and measure myelin loss and repair in MS lesions \cite{ANA24620}. The reader familiar with Alzheimer's disease (AD) may note that the tracer is the same as the one used in AD. PIB has the property to bind to proteins characterized by a similar conformation, contained in both amyloid plaques and myelin. Nevertheless, note that the signal in myelin is more subtle than for amyloid plaques. 
However, PET is expensive and not offered in the majority of medical centers in the world. Moreover, it is invasive due to the injection of a radioactive tracer. On the contrary, MR imaging is a widely available and non-invasive technique, but existing MRI sequences do not provide, to date, a reliable, specific, or direct marker of demyelination or remyelination. Therefore, it would be of considerable interest to be able to predict the PET-derived myelin content map from multimodal MRI.

In recent years, various methods for medical image enhancement and synthesis using deep neural networks (DNN) have been proposed, such as reconstruction of 7T-like T1-w MRI from 3T T1-w MRI \cite{Bahrami2016} and generation of FLAIR from T1-w MRI \cite{Sevetlidis2016}. Several works have proposed to predict PET images from MRI or CT images \cite{10.1007/978-3-319-67564-0_5,10.1007/978-3-319-68127-6_6,pmid25320813}. A single 2D GAN has been proposed to generate FDG PET from CT for tumor detection in lung \cite{10.1007/978-3-319-67564-0_5} and liver region \cite{10.1007/978-3-319-68127-6_6}. However, they do not take into account the spatial nature of 3D images and can cause discontinuous predictions between adjacent slices. Additionally, a single GAN can be difficult to train when the inputs become complex. A two-layer DNN has been proposed to predict FDG PET from T1-w MRI \cite{pmid25320813} for AD diagnosis. However, a 2-layer DNN is not powerful enough to fully incorporate the complex information from multimodal MRI. Importantly, all these works were devoted to the prediction of FDG PET. Predicting myelin information (as defined by PIB PET) is a more difficult task because the signal is more subtle and with weaker relationship to anatomical information that could be found in T1-w MRI or CT. 




In this work, we therefore propose Sketcher-Refiner GANs consisting in two conditional GANs (cGANs) with specifically designed adversarial loss to predict the PET-derived myelin content from multimodal MRI. Compared to previous works \cite{10.1007/978-3-319-67564-0_5,10.1007/978-3-319-68127-6_6} using GANs for medical imaging, our method solves the prediction problem by a sketch-refinement process which  decomposes the difficult problem into more tractable subproblems and makes the GANs training process more stable. In our method, the sketcher generates the preliminary anatomy and physiology information and the refiner refines and generates images reflecting the tissue myelin content in the human brain. In addition, the adaptive adversarial loss is designed to force the refiner to pay more attention to the prediction of the myelin content in MS lesions and normal appearing white matter (NAWM). To our knowledge, this is, to date, the first work to predict, from multi-sequence MR images, the myelin content measured usually by PET imaging. 



\section{Method}

\subsection{Sketcher-Refiner Generative Adversarial Networks (GANs)}
GANs are generative models consisting of two components: a generator G and a discriminator D. Given a real image $y$, the generator G aims to learn the mapping $\mathrm{G}(z)$ from a random noise vector $z$ to the real data. The discriminator $\mathrm{D}(y)$ evaluates the probability that $y$ is a true image. To constrain the output of the generator, cGANs \cite{MirzaO14} were proposed in which the generator and the discriminator both receive a conditional variable $x$. More precisely, D and G play the two-player conditional minimax game with the following cross-entropy loss function:
\begin{equation}
\begin{aligned}
\min_{G} \max_{D}\mathcal{L}(D,G)=\mathbb{E}_{
	x,y\sim{
	p_{\mathrm{data}}(x,y)
	}
}[\log{D}(x,y)]-\\
\mathbb{E}_{
	x\sim{
	p_{\mathrm{data}}(x)
	},
	z\sim{
	p_{\mathrm{z}}(z)
	}
}[\log(1-D(x,G(x,z)))]
\end{aligned}
\end{equation} 
where $p_\mathrm{data}$ and $p_\mathrm{z}$ are the distributions of real data and the input noise.
\vspace{-1em}
\begin{figure*}[ht!]
\includegraphics[scale=0.35]{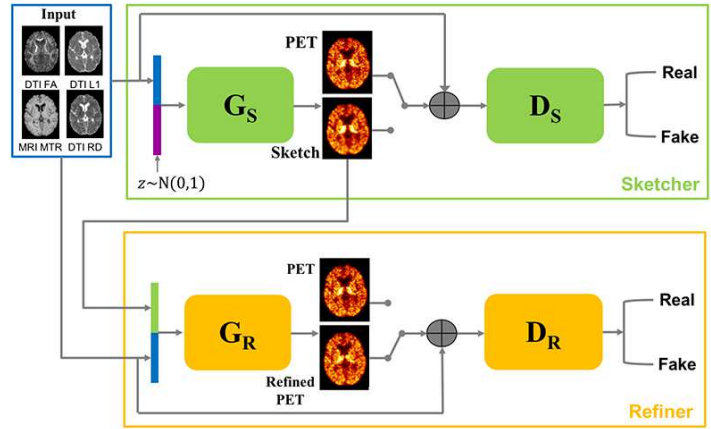}
\centering
\vspace{-1em}
\caption{The proposed sketcher-refiner GANs. The sketcher receives MR images and generates the preliminary anatomy and physiology information. The refiner receives MR images $I_{\mathrm{M}}$ and the sketch $I_{\mathrm{S}}$. Then it refines and generates PET images.}
\label{model}
\end{figure*}
\vspace{-1em}

The goal is to predict the $[^{11} \mbox{C}] \mbox{PIB}$ PET distribution volume ratio (DVR) parametric map $I_{\mathrm{P}}$ which reflects the myelin content, from multimodal MRI $I_{\mathrm{M}}$. Figure.~\ref{model} is the architecture of our method consisting of two cGANs named \textbf{Sketcher} and \textbf{Refiner}. Our method decomposes the prediction problem into two steps: 1) sketching anatomy and physiology information and 2) refining and generating images reflecting the tissue myelin content in the human brain. To the end, the sketcher and the refiner have the following cross-entropy losses:









\begin{equation}
\begin{aligned}
\min_{G_{\mathrm{S}}} \max_{D_{\mathrm{S}}}\mathcal{L}(D_{\mathrm{S}},G_{\mathrm{S}})=\mathbb{E}_{
	I_{\mathrm{M}},I_{\mathrm{P}}\sim{
	p_{\mathrm{data}}(I_{\mathrm{M}},I_{\mathrm{P}})
	}
}[\log{D_{\mathrm{S}}}(I_{\mathrm{M}},I_{\mathrm{P}})]-\\
\mathbb{E}_{
	I_{\mathrm{M}}\sim{
	p_{\mathrm{data}}(I_{\mathrm{M}})
	},
	z\sim{
	p_{\mathrm{z}}(z)
	}
}[\log(1-D_{\mathrm{S}}(I_{\mathrm{M}},G_{\mathrm{S}}(I_{\mathrm{M}},z)))]\\
\end{aligned}
\end{equation} 
\begin{equation}
\begin{aligned}
\min_{G_{\mathrm{R}}} \max_{D_{\mathrm{R}}}\mathcal{L}(D_{\mathrm{R}},G_{\mathrm{R}})=\mathbb{E}_{
	I_{\mathrm{M}},I_{\mathrm{P}}\sim{
	p_{\mathrm{data}}(I_{\mathrm{M}},I_{\mathrm{P}})
	}
}[\log{D_{\mathrm{R}}}(I_{\mathrm{M}},I_{\mathrm{P}})]-\\
\mathbb{E}_{
	I_{\mathrm{M}}\sim{
	p_{\mathrm{data}}(I_{\mathrm{M}})
	},
	I_{\mathrm{s}}\sim{
	G_{\mathrm{S}}(I_{\mathrm{M}},z)
	}
}[\log(1-D_{\mathrm{R}}(I_{\mathrm{M}},G_{\mathrm{R}}(I_{\mathrm{M}},I_{\mathrm{s}})))]
\end{aligned}
\end{equation} 
where $D_{\mathrm{S}}$, $D_{\mathrm{R}}$ and $G_{\mathrm{S}}$, $G_{\mathrm{R}}$ represent the discriminator and the generator in the sketcher and the refiner respectively. 
\vspace{-1em}

\subsection{Adversarial Loss with Adaptive Regularization} 
\label{sub:generator_with_}
Previous works \cite{pix2pix2016} have shown that it can be useful to combine the GAN objective function with a traditional constraint, such as L1 and L2 loss. They further suggested using L1 loss rather than L2 loss to encourage less blurring. We hence mixed the GANs' loss function with the following L1 loss for the sketcher.

\vspace{-1em}
\begin{equation}
\begin{aligned}
\mathcal{L}_{L1}(G_{\mathrm{S}})=\frac{1}{N}\sum_{i=1}^{N}|I_{\mathrm{P}}^{i}-G_{\mathrm{S}}(I_{\mathrm{M}}^{i},z^{i})|
\end{aligned}
\end{equation}
where $N$ is the number of subjects and $i$ denotes the index of a subject.

In CNS, myelin constitutes most of the white matter (WM). To force the generator on MS lesions where demyelination happens, the whole image is divided into three regions of interest (ROIs): lesions, NAWM and ``other". We thus defined for the refiner a weighted L1 loss in which the weights are adapted to the number of voxels in each ROI indicated as $N_{\mathrm{Les}}$, $N_{\mathrm{NAWM}}$ and $N_{\mathrm{other}}$. Given the masks of the three ROIs: $R_{\mathrm{Les}}$, $R_{\mathrm{NAWM}}$ and $R_{\mathrm{other}}$, the weighted L1 loss for the refiner is defined as follows:
\vspace{-1em}
\begin{equation}
\begin{aligned}
\mathcal{L}_{L1}(G_{\mathrm{R}})=\frac{1}{N\times{M}}\sum_{i=1}^{N}\Big(\frac{1}{N_{\mathrm{Les}}}\sum_{j\in{R_{\mathrm{Les}}}}|I_{\mathrm{P}}^{i,j}-\hat{I}_{\mathrm{P}}^{i,j}|+\\
\frac{1}{N_{\mathrm{NAWM}}}\sum_{j\in{R_{\mathrm{NAWM}}}}|I_{\mathrm{P}}^{i,j}-\hat{I}_{\mathrm{P}}^{i,j}|+
\frac{1}{N_{\mathrm{other}}}\sum_{j\in{R_{\mathrm{other}}}}|I_{\mathrm{P}}^{i,j}-\hat{I}_{\mathrm{P}}^{i,j}|
\Big)
\end{aligned}
\end{equation} 
where $M$ is the number of voxels in a PET image, $j$ is the index of a voxel and $\hat{I}_{\mathrm{P}}$ is the prediction output from the refiner. 

To sum up, our overall objective functions are defined as follows:

\begin{equation}
\begin{aligned}
G_{\mathrm{S}}^{*}=\arg\min_{G_{\mathrm{S}}} \max_{D_{\mathrm{S}}}\mathcal{L}(D_{\mathrm{S}},G_{\mathrm{S}})+\lambda_{\mathrm{S}}\mathcal{L}_{L1}(G_{\mathrm{S}}) \\
G_{\mathrm{R}}^{*}=\arg\min_{G_{\mathrm{R}}} \max_{D_{\mathrm{R}}}\mathcal{L}(D_{\mathrm{R}},G_{\mathrm{R}})+\lambda_{\mathrm{R}}\mathcal{L}_{L1}(G_{\mathrm{R}})
\end{aligned}
\end{equation}        
where $\lambda_{\mathrm{S}}$ and $\lambda_{\mathrm{R}}$ are hyper-parameters which balance the contributions of two terms in the sketcher and the refiner respectively.
\vspace{-1em}

\subsection{Network architectures} 
\label{sub:subsection_name}

\begin{figure*}[ht!]
\includegraphics[scale=0.35]{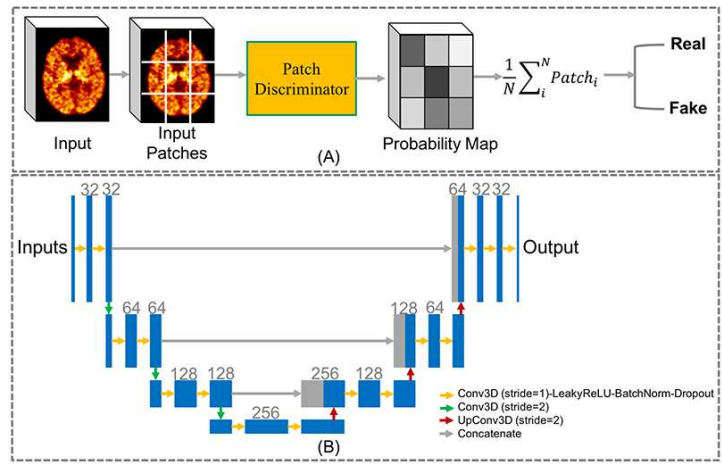}
\centering
\vspace{-1em}
\caption{The D and the G in our GANs. (A) The proposed 3D patch discriminator which takes all the patches and classifies them separately to output a final loss. (B) The 3D U-Net shaped generator with implementation details shown in the image.}
\label{patchdis}
\end{figure*}

Both the sketcher and the refiner have the same architectures for their generators and discriminators. For the generators, a general shape of a “U-Net” is used (see details in Fig.~\ref{patchdis} (B)). We use LeakyReLU which allows a stable training of GANs with 0.2 as slope coefficient. 

For the discriminator, a traditional approach in GANs is to use a global discriminator: the discriminator is trained to globally distinguish if the input comes from the true dataset or from the generator. However, when we trained, the generator tries to over-emphasize certain image features in some regions so that it can  make the global discriminator fail to differentiate a real or fake image. In our problem, each region in the PET image has its own myelin content. A key observation is that any local region in a generated image should have a myelin content that is similar to that of the real local region. Therefore, instead of using a traditional global network, we define a 3D patch discriminator trained by local patches from input images. As shown in Fig.~\ref{patchdis} (A), the input image is firstly divided into patches with size $l \times w \times h$ and then the 3D patch discriminator classifies all the patches separately. The final loss of the 3D patch discriminator is the sum of the cross-entropy losses from all the local patches. Its architecture is a traditional CNN including four stages of Conv3D-BatchNorm-LeakyReLU-Downsampling and a combination with a fully connected layer and a Softmax layer as last two layers.
\vspace{-2em}
\vspace{-3em}


\section{Experiments and Evaluations} 
\label{sec:evaluations}

\subsection{Overview}
\begin{itemize}

\item[-] \textbf{Dataset:} Our dataset includes 18 MS patients (12 women, mean age 31.4 years, sd 5.6) and 10 age- and gender-matched healthy volunteers (8 women, mean age 29.4, sd 6.3). For each subject, the MRI data includes a magnetisation transfer ratio map (MTR) and three measures derived from diffusion tensor imaging (DTI): fractional anisotropy (FA), radial diffusivity (RD), axial diffusivity (AD) while the PET data is a  $[^{11} \mbox{C}] \mbox{PIB}$ PET DVR parametric map. The preprocessing steps mainly consist of the intra-subject registration onto $[^{11} \mbox{C}] \mbox{PIB}$ PET image space.
\item[-] \textbf{Training details:} Our sketcher-refiner GANs is implemented with the Keras library. The convolution kernel size is $3 \times 3 \times 3$ and the rate for dropout layer is 50\%. The optimization is performed with the ADAM solver with $10^{-4}$, $5\times{10^{-5}}$ as initial learning rates for the sketcher and the refiner respectively. We used 3-fold cross validation with 19 subjects for training and 9 subjects for testing. Two GTX 1080 Ti GPUs are used for training.

\end{itemize}

\vspace{-2em}

\subsection{Qualitative Evaluation}

\begin{figure*}[ht!]
\includegraphics[scale=0.34]{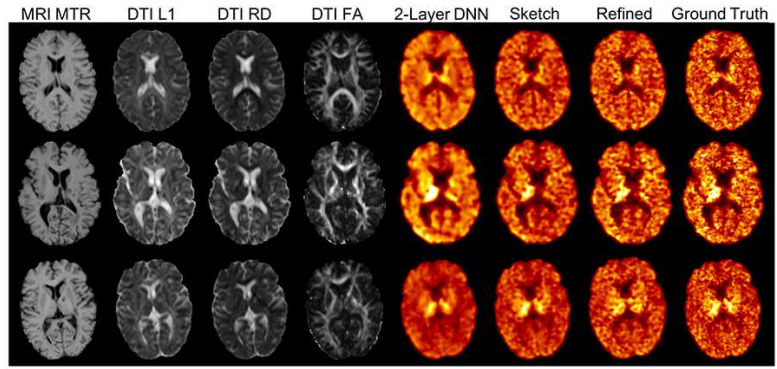}
\centering
\vspace{-1em}
\caption{Qualitative comparison of our method (``Refined"), a 2-layer DNN and the single cGAN (corresponding to the sketcher in our approach is denoted as ``Sketch") as well as the ground truth and the input MR images.
}
\label{sample}
\end{figure*}

Figure.~\ref{sample} shows the qualitative comparison of our prediction results, a 2-layer DNN as in \cite{pmid25320813} and a single cGAN (corresponding to the sketcher in our approach) with corresponding input multimodal MRI and the true $[^{11} \mbox{C}] \mbox{PIB}$ PET DVR parametric map. We can find that the 2-layer DNN failed to find the non-linear mapping between the multimodal MRI and the myelin content in PET. Especially, some anatomical or structural traces (that are not present in the ground truth) can still be found in the 2-layer-DNN predicted PET. This highlights that the relationship between myelin content and multimodal MRI data is complex, and only two layers  are not powerful enough to encode-decode it.


It is also shown that the single-GAN predicted PET (sketch) generates a blurry output with the primitive shape and basic information. On the other hand, after the refinement process by our refiner, the output is closer to the ground truth and the myelin content is better predicted. According to this, we can also conclude that only using a single cGAN (sketcher) like in \cite{10.1007/978-3-319-67564-0_5,10.1007/978-3-319-68127-6_6} is insufficient for our problem. 

\vspace{-2em}
\vspace{-3em}

\subsection{Global Evaluation of Myelin Prediction}
\begin{figure*}[ht!]
\includegraphics[scale=0.34]{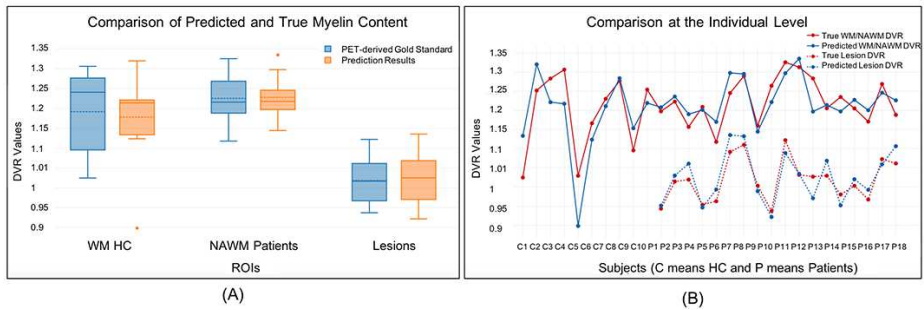}
\centering
\vspace{-2em}
\caption{
(A) Group level evaluation. The box plots show the median (middle solid line), mean (middle dotted line) and the range of DVR for each ROI for PET-derived DVR parametric map used as gold standard (blue) and the prediction results (yellow). (B) Individual level evaluation. The red and blue colors respectively represent the mean DVR values for each subject extracted in the ground truth image and the predicted PET, respectively. Results from WM/NAWM and lesions are shown using respectively a solid line and a dotted line.
}
\label{global}
\end{figure*}

To assess myelin content at a global level, three ROIs were defined: 1) WM in healthy controls (HC); 2) NAWM in MS patients; 3) lesions in MS patients. Figure~\ref{global}(A) displays the comparison between the distribution of the ground truth and the predicted PET. It shows that the PET-derived overall patterns can be well reproduced by the synthetic data. Specifically, both with the gold standard and the synthetic data, there is no significant difference between NAWM in patients and WM in HC, while a statistically significant reduction of myelin content  in lesions compared to NAWM can be found ($p<0.0001$). Figure~\ref{global}(B) shows the comparison at the individual participant level. The predicted DVR map is very close to the ground truth in the vast majority of participants. This demonstrates that our method can adequately predict myelin content at the individual patient level.

\vspace{-2em}
\vspace{-3em}

\subsection{Voxel-wise Evaluation of Myelin Prediction}
\begin{figure*}[ht!]
\includegraphics[scale=0.22]{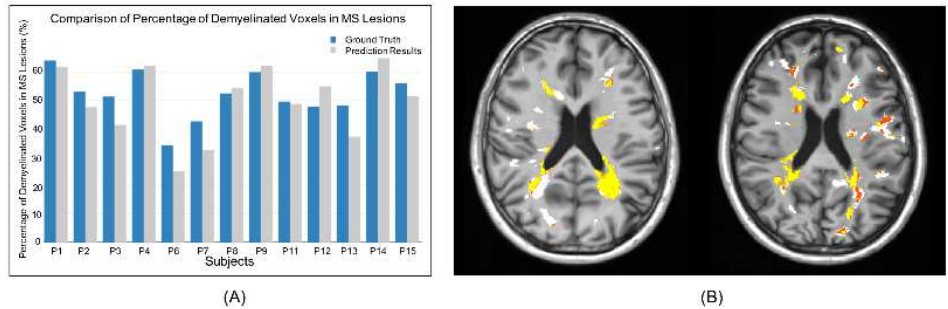}
\centering
\vspace{-2em}
\caption{(A) Percentage of demyelinated voxels in white matter MS lesions for each patient computed from the ground truth (blue) and from our method (grey). (B) Examples of demyelinated voxels classified from the ground truth and our predicted PET within MS lesions. Agreement between methods is marked in yellow (both true and predicted PET indicated demyelination) and white (both methods did not indicate demyelination). Disagreement is marked in red (demyelination only with the true PET) and orange (only with the predicted PET).
}
\label{demy}
\end{figure*}

We also evaluate the ability of our method to predict myelin content at the voxel-wise level. Within each MS lesion of each patient, each voxel was classified as a demyelinated voxel according to a procedure defined in a previous clinical study \cite{ANA24620}. We hence measure the percentage of demyelinated voxels over total lesions load of each patient for both the ground truth and the predicted PET as shown in Figure~\ref{demy} (A). Our prediction results approximate the ground truth for most of the patients. The average Dice index between the demyelination map derived from the ground truth and  our predicted PET is 0.83. Figure~\ref{demy} (B) shows demyelinated voxels classified from both the true and the predicted PET within MS lesions. The large agreement regions demonstrate our method's strong ability to predict the demyelination in MS lesions at the voxel-wise level. 


\vspace{-1em}
\section{Conclusion and Future Work} 
\label{sec:discussion}
We proposed Sketcher-Refiner GANs with specific designed adversarial loss functions to predict the PET-derived myelin content from multimodal MRI. The prediction problem is solved by a sketch-refinement process in which the sketcher generates the preliminary anatomy and physiology information and the refiner refines and generates images reflecting the tissue myelin content in the human brain. Our method is evaluated for myelin content prediction at both global and voxel-wise levels. The evaluation results show that the demyelination in MS lesions, and myelin content in both patient's NAWM and control's WM can be well predicted by our method. In the future, it would be interesting to use our method on longitudinal dataset, to investigate dynamic demyelination and remyelination processes.


%
\vspace{-1em}
\bibliographystyle{splncs}
\bibliography{paper}

\end{document}